\documentclass[sigconf, nonacm]{acmart}
\AtBeginDocument{%
  }

\usepackage{balance} % for balancing columns on the final page
\usepackage{graphicx}
\usepackage{subcaption}
\usepackage{mdframed}
\usepackage{adjustbox}
\usepackage{booktabs}
\usepackage{multirow}

\newcommand\blfootnote[1]{%
  \begingroup
  \renewcommand\thefootnote{}\footnote{#1}%
  \addtocounter{footnote}{-1}%
  \endgroup
}

\setcopyright{acmlicensed}
\copyrightyear{2018}
\acmYear{2018}
\acmDOI{XXXXXXX.XXXXXXX}
\acmConference[ICAIL'26]{21st International Conference on Artificial Intelligence and Law}{June 08--12,
  2026}{Singapore}
\acmISBN{978-1-4503-XXXX-X/2018/06}

\begin{document}

\title[Confronting Label Indeterminacy in Automated Bail Decisions]{Confronting Label Indeterminacy in Automated Bail Decisions}

\author{Cor Steging}
\email{c.c.steging@rug.nl}
\orcid{0000-0001-6887-1687}
\affiliation{%
  \institution{Bernoulli Institute of Mathematics, Computer Science and Artificial Intelligence, University of Groningen}
  \country{The Netherlands}
}

\author{Tadeusz Zbiegień}
\email{tadeusz.zbiegien@doctoral.uj.edu.pl}
\orcid{0000-0001-9052-6978}
\affiliation{%
  \institution{Department of Legal Theory, Jagiellonian University}
  \country{Poland}
}

% \author{Contribution ID: 123}

% \renewcommand{\shortauthors}{Contribution ID: 123}
\renewcommand{\shortauthors}{Steging \& Zbiegień}

\acmSubmissionID{TBD}

\begin{abstract}
Bail decisions present a fundamental challenge for data-driven decision support systems. When bail is denied, the counterfactual outcome of whether the defendant would have appeared in court remains unobserved. 
As a result, historical bail data embed structural label indeterminacy: future decisions are influenced by past decisions whose outcomes are only partially knowable. 
Building automated systems on such data risks introducing bias and reinforcing feedback loops.
This raises a core question for machine-learning systems intended to assist judicial actors: how should cases in which bail was denied be treated during model development? 
In a case study of bail decisions from the Unified Judicial System of Pennsylvania, we evaluate five contemporary approaches to handling label indeterminacy across three machine learning models, including a novel label imputation method motivated by the dynamics of bail decisions.
Each method relies on unverifiable assumptions, yet all influence the models’ predictive behaviour, sometimes even more so than the choice of model itself.
Explainable AI analysis further reveals that these effects extend to the models’ internal decision-making processes as well.
Finally, we consider the notion of label indeterminacy from a legal perspective and assess the legitimacy of these approaches in the context of bail decision-making.
\end{abstract}

\keywords{Label indeterminacy, Machine Learning, Bail decisions}


\maketitle
\blfootnote{This manuscript has been accepted for presentation as a short paper at the 21st International Conference of AI \& Law in Singapore, June 8 to 12 of 2026.}

\section{Introduction}
In the United States, more than two-thirds of the jail population consists of individuals who are legally presumed innocent and held in pretrial detention, a figure that continues to grow and imposes substantial social and financial costs on defendants and taxpayers alike~\cite{RANSON2023102872}. Decisions about whether to grant bail or to detain a defendant pretrial therefore constitute a central challenge in the criminal justice system.

Technological decision support systems have been introduced to make bail decision-making more efficient and, potentially, more rational and thus equitable~\cite{AbuElyounes2020BailOrJail}. These systems typically aim to predict the likelihood that a defendant will fail to appear in court or whether they are likely to pose a risk to public safety, thereby informing judicial bail decisions.
A well-known, yet controversial example of computer programs that assist in bail predictions is COMPAS, a rule-based system originally designed to predict recidivism~\cite{COMPAS}. 
More recently, it has been argued that modern AI solutions may contribute to enhancing judges’ actual, rich, and perceived trustworthiness in the context of bail decision-making~\cite{Morin-Martel2024}.

At the same time, automated bail decision systems face significant challenges: models can overestimate risk due to reliance on outdated data, embed implicit moral bias, and project a misleading appearance of scientific objectivity that may normalize preventive detention~\cite{koepke2018danger}. 
In practice, this can result in systematically inflated risk estimates, unexamined normative assumptions regarding how risk scores are used, and increased judicial acceptance of preventive detention.

\begin{figure*}
    \includegraphics[width=0.9\linewidth]{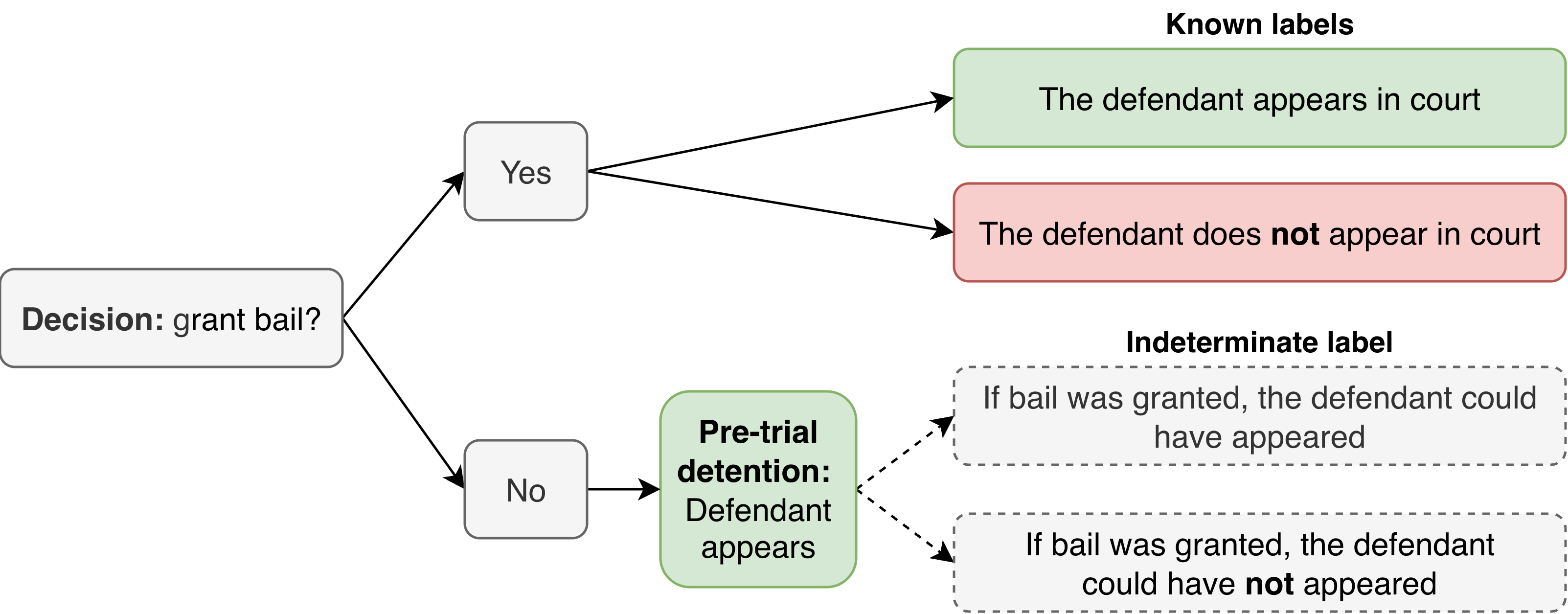}
    \caption{Flowchart of the bail decision process illustrating how missing counterfactuals give rise to indeterminate labels.}
    \label{fig:bail_figure}
\end{figure*}

Furthermore, comparing automated decisions with human judgment is inherently difficult due to the absence of counterfactual data~\cite{Kleinberg2017-xj}. As illustrated in Figure~\ref{fig:bail_figure}, when a defendant is denied bail and detained pretrial, they will in practice always appear in court, leaving it unknown whether they would have appeared or posed a risk had bail been granted. Consequently, automated predictions in these cases can only be compared to human decisions, rather than evaluated against a normative ground truth of actual risk or appearance. 
Moreover, when training an automated AI system on historical bail cases, the absence of counterfactual information renders the labels of pretrial detention cases indeterminate, as court appearance is observed only as a consequence of the detention decision rather than the defendant’s true underlying propensity to appear (see Figure~\ref{fig:bail_figure}). While the problem of counterfactuals is not new and prior work has shown that using such data without accounting for indeterminate labels can perpetuate historical bias~\cite{COMPAS}, the challenge of handling these indeterminate labels in a machine learning context has yet to be addressed. With automated decision support systems increasingly influencing bail outcomes, properly addressing this issue is critical.

Methods exist to impute indeterminate labels, but all depend on unverifiable assumptions~\cite{Schoeffer2025Perils}. Prior work on label indeterminacy in the legal domain has focused on appeals, showing that different approaches for handling indeterminate labels can produce substantial variation in model behaviour~\cite{StegingJURIX2025}. In this study, we examine label indeterminacy in the context of bail decisions. We apply five label imputation methods to machine learning models trained to predict failure to appear and evaluate these methods from a legal perspective. %to assess their practical feasibility.

\section{Background}
%General information about bails. 
%How they work
%An issue not only in US but in general

%An increasing number of U.S. states are eliminating cash bail, which has been shown to %disproportionately disadvantage defendants with limited financial resources, and are instead adopting risk assessment systems~{Citationneeded}. 
%These systems, including those used in jurisdictions such as Washington, D.C., rely on algorithmic models that incorporate defendants’ criminal histories and other personal characteristics~{Citationneeded}.

%Previous work on AI \& Law with bail:
%Using LLMs and RAG to predict bail in India~\cite{Srivastava2025IBPSIB}.
%IndicBERT, a language model pre-trained in Indian languages, which is fine-tuned on the HLDC dataset to handle bail prediction tasks~\cite{Zacharia2025legalinsight}.
%Survey about the use of ML in automated bail decisions~\cite{Morin-Martel2024}.

%Predicting recidivism:
%COMPAS predicts recidivism and was used for bail decisions~\cite{COMPAS}.
%Department of justice did recidivism forecasting using XGBoost~\cite{Han2021RecidivismXGBoost}
%

Bail is usually understood as a conditional release of the accused before trial, allowing them to remain free pending court proceedings, usually on the condition that they agree to appear in court and sometimes also on the condition that they provide financial security. It aims to balance the presumption of innocence, personal liberty, public safety, and the obligation to appear in court \cite{AbuElyounes2020BailOrJail}. A growing number of US states are eliminating cash bail, as it has been argued that it can disproportionately discriminate against defendants with limited financial resources, and are instead introducing risk assessment systems \cite{AbuElyounes2020BailOrJail}. These systems rely on algorithmic models that take into account defendants' criminal history and other personal characteristics. However, the issue of pretrial assessment of defendants and their possible detention generates controversial problems that are faced not only by the United States but also by many legal systems around the world. %In many countries, bail is treated as an exception, and courts prefer release without financial conditions. 

In the field of artificial intelligence and law, previous work has explored the use of artificial intelligence to support bail decisions~\cite{Morin-Martel2024}.
~\cite{AbuElyounes2020BailOrJail} provides an analysis of pretrial risk assessment tools highlighting the arguments raised by both proponents and opponents. Research on New York City cases from 2008 to 2013 suggested potential improvements in bail decision-making~\cite{Kleinberg2017-xj}, whereas a large randomized field trial found that providing judges with Public Safety Assessment (PSA) risk scores had little impact on detention outcomes~\cite{Imai2023AlgorithmAssistedDecision}.Related research has focused on predicting recidivism, which often informs pretrial risk assessment. The COMPAS tool has been used in the US for bail decisions~\cite{COMPAS}, and the Department of Justice has reported on algorithmic approaches to recidivism prediction as well~\cite{Han2021RecidivismXGBoost}.

AI for automated bail decisions has also been explored in other jurisdictions.
In India, fine-tuned language models~\cite{Zacharia2025legalinsight} and large language models combined with retrieval-augmented generation (RAG)~\cite{Srivastava2025IBPSIB} have been applied to predict bail outcomes. In the UK, fairness-aware machine learning models have been shown to predict general and violent recidivism with high accuracy~\cite{Verrey2025FairnessScaleRecidivism}, while in Australia predictive modeling shows potential to support bail decisions but raises complex jurisprudential challenges~\cite{Hansard_Zhou_2025}. 

Other work has examined the ethical and normative dimensions of AI in bail, including different conceptions of moral responsibility for AI and human agents~\cite{Lima2021}. Public attitudes toward AI-assisted judges have been shown to vary across racial groups, with judges relying on their own expertise generally rated more favorably than those relying on AI support~\cite{Fine2025PublicPerceptionsAI}.

The issue of counterfactuals is unavoidable in bail decision-making, since only one potential outcome is ever observed, with substantial implications for modeling and downstream applications~\cite{Kleinberg2017-xj}. Prior work has addressed this challenge by studying fairness in risk assessment tools under counterfactual settings and developing predictors that aim to ensure an equitable distribution of benefits and harms
% , an approach applied, among others, to the COMPAS dataset
~\cite{mishler2021}.
Other work introduces the concept of counterfactual loss to assess decisions based on all possible outcomes~\cite{koch2025statisticaldecisiontheorycounterfactual}. 
We, in turn, tackle the problem from the perspective of label indeterminancy.  

Previous research on label indeterminacy has examined survival prediction in the medical domain, where decisions about withholding life-sustaining treatment are considered~\cite{Schoeffer2025Perils}, and in the legal domain, focusing on overturned decisions in the European Court of Human Rights~\cite{StegingJURIX2025}. In both contexts, identifying indeterminate labels requires domain expertise, and methods have been explored and developed to impute these labels. Although all methods rely on unverifiable assumptions, differences in model behaviour are consistently observed depending on the imputation method employed.

\section{Method}
To investigate the effects of label indeterminacy in bail decisions, we train three machine learning models to predict defendant appearance while accounting for indeterminate cases where the outcome was determined by pretrial detention using five different label imputation methods. We apply each method and examine how it influences model behaviour. In this section, we describe the dataset, define label indeterminacy in the domain, outline the methods for handling indeterminate labels, and present the experimental setup.

\subsection{Dataset}
The dataset that we use consists of 90,732 publicly available and anonymized court cases from the Unified Judicial System of Pennsylvania, spanning January 2016 to June 2020~\cite{williams2021bayesian}. It is a tabular dataset that includes personal information about defendants, such as age and gender, as well as their criminal history, including the number of felonies and misdemeanors. Case-specific information is also recorded, including the county of the court, the type of attorney, the type of bail requested (e.g., monetary), and the status of the bail request (e.g., denied).

The target feature in this dataset is whether a defendant appears in court or fails to appear (FTA). We predict this outcome by training machine learning models on the remaining features. In principle, these predictions could provide judges with additional information when deciding whether to grant bail.

Label indeterminacy is central to this task. When bail is denied, defendants almost always appear in court because they are held in pretrial detention. In our dataset, only 0.01\% of bail-denied cases result in a defendant failing to appear, typically due to extreme events such as escape or death. This means that nearly all bail-denied cases produce a negative FTA label. However, this does not indicate whether the defendant would have appeared if bail had been granted, rendering these labels indeterminate. Machine learning models trained on such cases must therefore account for this indeterminacy.

Labels for denied cases are unambiguously indeterminate, but we classify several other case types as indeterminate as well.
We define a label as indeterminate when an intervention alters the observed outcome such that it no longer reliably reflects the defendant’s propensity to fail to appear. In all such cases, the defendant was detained pretrial, meaning the counterfactual outcome under release is unobserved.

On this basis, we also classify cases with bail status \textit{Set} or \textit{Partial Posting} as indeterminate. In these cases, defendants have not paid, or have only partially paid, their monetary bail, which typically results in pretrial detention and subsequently a court appearance in 99.3\% of cases. The observed FTA outcomes thus reflect detention rather than the defendant’s underlying willingness to appear.
Similarly, the \textit{Bond Terminated} status encompasses multiple permissible interventions (e.g., case resolution, dismissal, or transfer) that are not directly linked to appearance behaviour, yet typically result in court appearance due to pretrial detention (93.81\%).

Conversely, we treat cases with the bail status \textit{Posted}, indicating that bail has been paid or that alternative release conditions have been satisfied, as determinate. In these cases, defendants were released pretrial, and no intervention directly enforces court appearance or prevents failure to appear.
Cases with the status \textit{Forfeited} are likewise considered determinate. These cases involve defendants who were granted bail but failed to appear at a scheduled hearing, resulting in bail forfeiture and subsequent pretrial detention. The failure to appear is therefore directly observed.
The same applies to cases with the status \textit{Revoked} when revocation is explicitly due to a failure to appear. In contrast, revoked cases based on violations unrelated to failure to appear are treated as indeterminate, as bail loss in these instances is not causally attributable to non-appearance but instead arises from pretrial detention.
Table~\ref{tbl:indeterminacy_overview} summarizes the determinate and indeterminate labels.

% Maybe a table here?
\begin{table}[t]
\centering
\caption{Label indeterminacy in the dataset.}
\begin{tabular}{ll}
\textbf{} & \textbf{Bail Status} \\
% \midrule
\toprule
\textbf{Determinate} 
    & Posted \\
    & Forfeited \\
    & Revoked (due to FTA) \\
\midrule
\textbf{Indeterminate}
    & Denied \\
    & Set \\
    & Partial Posting \\
    & Bond Terminated \\
    & Revoked (other reasons) \\
\bottomrule
\end{tabular}
\label{tbl:indeterminacy_overview}
\end{table}

\subsection{Preprocessing}
To prepare the data for our experiments, we randomly sample 20\% of the data, stratified by the FTA label, and set these aside as our test set. As the cases span 2016–2020, we assume temporal effects to be negligible.
The test set consists of 18,147 cases, and the remaining 72,587 cases form the training set. In both datasets, approximately 3.8\% of defendants failed to appear.

For training, we require a balanced label distribution, with 50\% non-FTA and 50\% FTA cases. As only 3.8\% of the data involve FTA, we apply deterministic undersampling with full minority reuse to construct 25 balanced training sets. The FTA cases are identical across sets, while the non-FTA cases do not overlap. We will train a different model on each of these subsets and average the performance across the 25 subsets.

We remove six features from the data: the date, magistrate, bailType, bailAmount, bailStatus, and bailDenied. The first two are removed as they should be irrelevant to predicting the FTA label. The latter four are removed as they are information that is only available after bail has already been granted. As the predicted FTA likelihood of our models should be used to inform bail decisions, we cannot include these features. We decided to include all other features, including sensitive attributes such as race and sex, to faithfully reflect the information present in the data. Furthermore, exclusion would not remove their influence but instead shift it to correlated proxies, complicating both interpretation and bias assessment.
We preprocess the data by midpoint-encoding binned numeric features such as age, ordinal-encoding ordered features like local employment levels, and one-hot encoding categorical variables, including race, sex, and type of attorney employed.

\subsection{Imputing indeterminate labels}

\begin{table*}[ht]
\centering
% \scriptsize
\caption{Overview of the label imputation methods used to account for indeterminate labels in the bail domain.}
\setlength{\tabcolsep}{3pt} % horizontal padding
\renewcommand{\arraystretch}{1.1} % vertical padding
% \begin{tabular}{p{0.07\textwidth} p{0.15\textwidth} p{0.29\textwidth} p{0.31\textwidth}}
\begin{tabular}{p{0.07\textwidth} p{0.14\textwidth} p{0.36\textwidth} p{0.36\textwidth}}
\toprule
\textbf{ID} & \textbf{Name} & \textbf{Description} & \textbf{Assumption } \\
\midrule
\texttt{$corr$} & Correct labels & Includes indeterminate cases without altering their labels. & Assumes defendants who were in pretrial detention would have appeared in court if they were not detained.  \\
\addlinespace[1mm]
\texttt{$daf$} & Detention-as-failure & Includes indeterminate cases and sets their label to 'fail to appear'. & Assumes defendants who were in pretrial detention would not have appeared in court if they were not detained.   \\
\addlinespace[1mm]
\texttt{$obs$} & Observed only & Includes only cases where bail was not denied and the outcome is observable. & Assumes missing at random; fails if the choice for the denial of bail is based on the content of the case. \\
\addlinespace[1mm]
\texttt{$obs+ip$} & Observed + IP & Trains only on non-denied cases and applies inverse propensity weights to correct for sample bias & Assumes that every case has a non-zero chance of being granted bail (positivity) and that all factors influencing whether bail is granted are captured in the data (no unobserved confounders) \\
\addlinespace[1mm]
\texttt{$nn$} & Nearest neighbor & Imputes the labels of bail-denied cases using the most similar cases from the bail-granted sample & Assumes there is a valid similarity metric between bail-denied and bail-granted cases. \\
\bottomrule
\end{tabular}
\label{tbl:methods}
\end{table*}

We examine five methods for imputing labels in indeterminate cases, as summarized in Table~\ref{tbl:methods} and described in detail in this section.
Each method is based on a defensible intuition but relies on unverifiable assumptions, so no single approach can be considered definitively correct.
Four of the methods are based on previous work~\cite{Schoeffer2025Perils, StegingJURIX2025}, while the $daf$ approach is a novel method specifically motivated by the dynamics of bail decisions.
We exclude approaches that require additional expert labeling, as used in other studies on label indeterminacy, to focus on methods that can be applied directly to existing datasets.

In the \textit{Correct labels} ($corr$) approach, we use the training data as is with the original labels unchanged, implicitly assuming that defendants held in pretrial detention would have appeared in court had they been released.

In the \textit{Detention-as-failure} ($daf$) approach, we also use the full training dataset but deterministically assign a failure-to-appear label to all indeterminate cases involving pretrial detention. This assumes that the Magisterial District Judge’s decision to order detention reflects an accurate expectation of non-appearance, and that defendants held pretrial would not have appeared in court had they been released. 

The \textit{Observed only} ($obs$) approach excludes all indeterminate cases, and trains models only on the determinate cases, as they have an observed label. This assumes that the indeterminate cases are a random sample of all cases.

The \textit{Observed + IP} ($obs\_ip$) approach corrects for sample bias from only observed labels by weighting each determinate case inversely to its probability of being observed, estimated via a propensity model (logistic regression) trained to distinguish determinate from indeterminate cases. This method assumes that the propensity model is correct and that all factors influencing bail decisions are captured in the dataset.

The \textit{Nearest neighbor} ($nn$) method does not disregard the indeterminate cases, but instead imputes their FTA label based on similar determinate cases using a nearest neighbor algorithm with $k=52$. This approach assumes that there is a valid similarity metric between cases, however.

As shown in Table~\ref{tbl:methods}, all methods rely on unverifiable assumptions, meaning that none can be considered definitively correct. In our experiments, we compare these methods and examine how they affect model behaviour.

\subsection{Experimental setup}
We evaluate the predictive behaviour of machine learning models trained to predict FTA based on case information. Three models are considered, all suitable for tabular bail decision data. First, we use logistic regression as a standard baseline. Second, following prior work on label indeterminacy in the medical domain, we include a random forest model~\cite{Schoeffer2025Perils}. Finally, we employ an XGBoost model, a widely used approach for tabular data that has also been applied to recidivism prediction~\cite{Han2021RecidivismXGBoost}. Full model parameters are provided in the Appendix (Table~\ref{tbl:appendix_parameters}).

Each model is trained on all 25 training subsets for each of the five label imputation methods. This results in a total of 348 trained models, for which we record predicted FTA probabilities on all cases in the test set.

% The nearest neighbor (NN) method depends on the choice of $k$, the number of neighbors. A common heuristic is to set $k=\sqrt{N}$, where $N$ is the number of training instances. In our subsets, the average number of cases with a determinate label is $N=2664.4$, yielding $\sqrt{2664.4}\approx 51.6$. We therefore set $k=51$.

%what about sort of bounding as in sensitivity analysis - e.g. two datasets; one with all denided cases FTA = 0; second with all denied cases FTA = 1; we train the same model separately on each dataset; we report the range of predictions -> we do not try to infer true labels, simply assessing sensitivity 

% or some sort of joint modeling - the bail decision and the FTA outcome ; instead of modeling FTA alone, we also explicitly model the decision; so for each case two variables - a decision variable (granted or denied) and an outcome variable on FTA which is observed only when bail is granted; when bail is denied FTA is unobserved and treated as latent variable

\section{Results}

\begin{table*}[t]
\caption{Mean MCC and standard deviation for each model across the 25 training sets on determinate and indeterminate cases of the test set. These values should be interpreted with caution: they both reflect performance on only a restricted, unrepresentative subset of the data, and the true labels of the indeterminate cases are considered indeterminate.}
\label{tbl:performances}
\begin{tabular}{rccccc}
\toprule
Method & $corr$ & $daf$ & $obs$ & $obs+ip$ & $nn$ \\
\midrule
\multicolumn{6}{l}{\textit{Determinate cases}} \\
\midrule
Logistic Regression & 10.78 $\pm$ 0.44 & 10.42 $\pm$ 0.23 & 10.85 $\pm$ 0.21 & 10.89 $\pm$ 0.22 & 10.70 $\pm$ 0.20 \\
Random Forest       & 11.37 $\pm$ 0.31 & 9.79  $\pm$ 0.26 & 11.13 $\pm$ 0.17 & 11.06 $\pm$ 0.21 & 10.53 $\pm$ 0.27 \\
XGBoost             & 10.61 $\pm$ 0.56 & 10.56 $\pm$ 0.30 & 11.41 $\pm$ 0.33 & 11.45 $\pm$ 0.31 & 10.99 $\pm$ 0.32 \\
\midrule
\multicolumn{6}{l}{\textit{Indeterminate cases}} \\
\midrule
Logistic Regression & 1.25 $\pm$ 0.58 & 1.00 $\pm$ 0.52 & 0.24 $\pm$ 0.50 & 0.60 $\pm$ 0.62 & 1.10 $\pm$ 0.59 \\
Random Forest       & 2.51 $\pm$ 0.53 & 1.53 $\pm$ 0.16 & 1.73 $\pm$ 0.34 & 1.87 $\pm$ 0.43 & 1.74 $\pm$ 0.30 \\
XGBoost             & 2.78 $\pm$ 0.52 & 1.52 $\pm$ 0.52 & 1.23 $\pm$ 0.61 & 1.15 $\pm$ 0.61 & 1.59 $\pm$ 0.73 \\
\bottomrule
\vspace{5pt}
\end{tabular}
\end{table*}

% \begin{figure}[t]
%     \centering
%     \begin{subfigure}[b]{0.5\textwidth}
%         \includegraphics[width=\textwidth]{figures/violin_plots_lr.png}
%         %\vspace{2pt}
%         \caption{Logistic Regression}
%         \label{fig:violin_lr}
%     \end{subfigure}
%     % \hfill
%     \vspace{2pt}
%     \begin{subfigure}[b]{0.5\textwidth}
%         \includegraphics[width=\textwidth]{figures/violin_plots_rf.png}
%         %\vspace{2pt}
%         \caption{Random Forest}
%         \label{fig:violin_rf}
%     \end{subfigure}
%     % \hfill
%     \vspace{2pt}
%     \begin{subfigure}[b]{0.5\textwidth}
%         \includegraphics[width=\textwidth]{figures/violin_plots_xgb.png}
%         %\vspace{2pt}
%         \caption{XGBoost}
%         \label{fig:violin_xgb}
%     \end{subfigure}
%     \caption{Violin plots displaying the distribution of the predictions for the three models trained using different label imputation methods.}
%     \label{fig:violins}
% \end{figure}

\begin{figure*}[t]
    \centering
    \begin{subfigure}[b]{0.75\textwidth}
        \includegraphics[width=\textwidth]{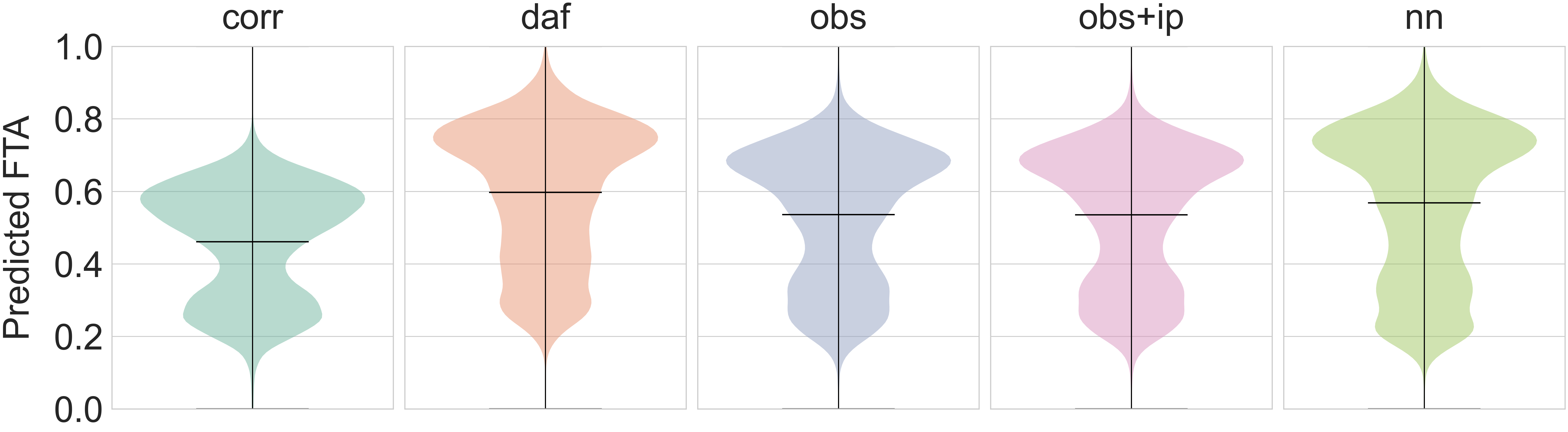}
        %\vspace{2pt}
        \caption{Logistic Regression}
        \label{fig:violin_lr}
    \end{subfigure}
    % \hfill
    \vspace{3pt}
    \begin{subfigure}[b]{0.75\textwidth}
        \includegraphics[width=\textwidth]{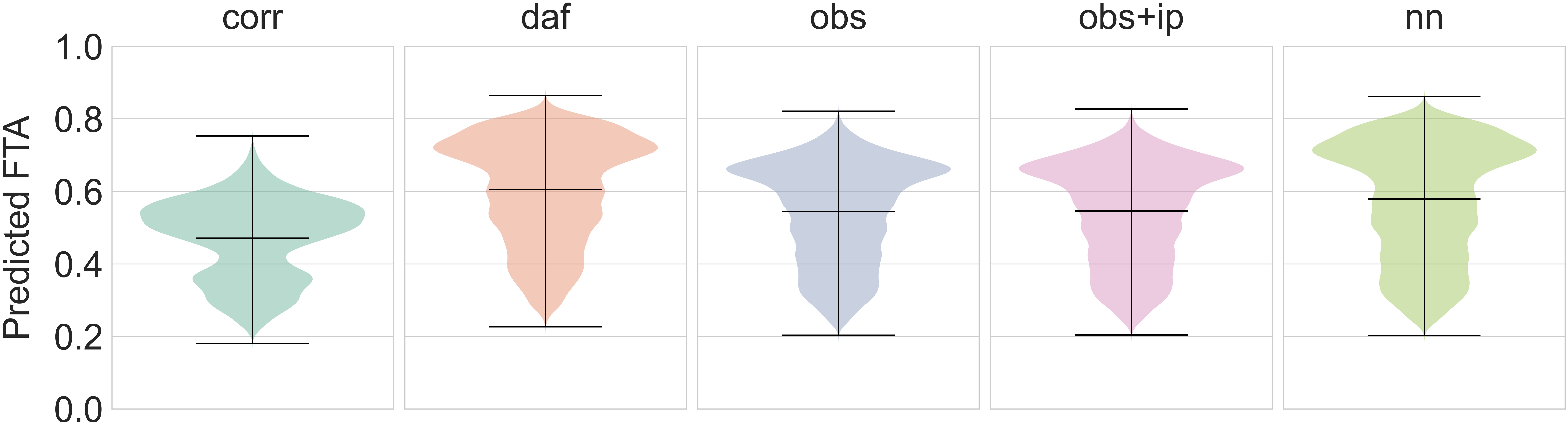}
        %\vspace{2pt}
        \caption{Random Forest}
        \label{fig:violin_rf}
    \end{subfigure}
    % \hfill
    \vspace{3pt}
    \begin{subfigure}[b]{0.75\textwidth}
        \includegraphics[width=\textwidth]{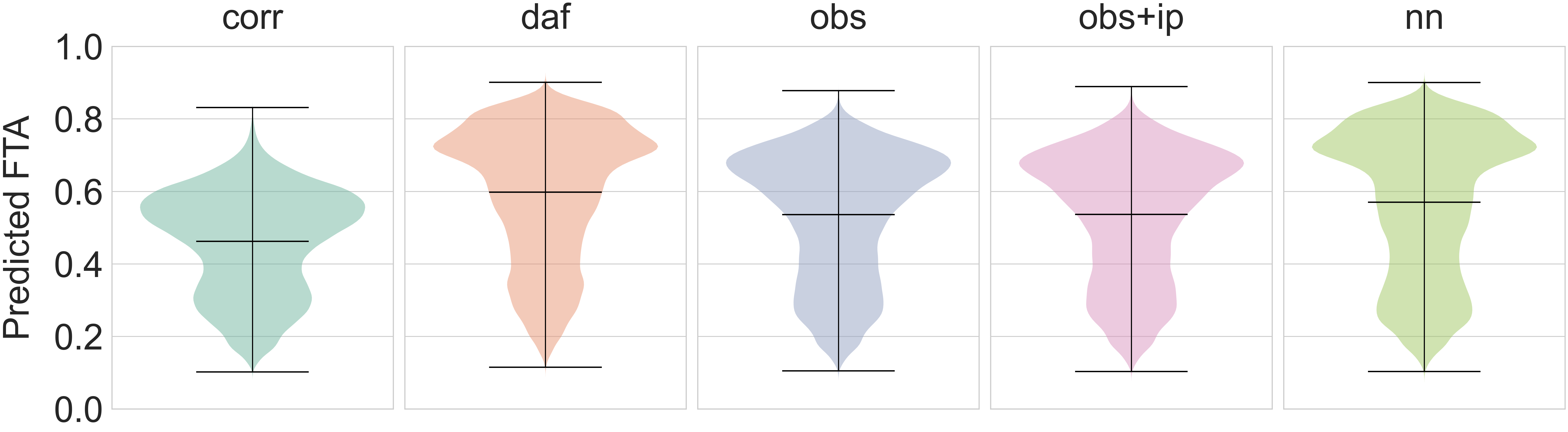}
        %\vspace{2pt}
        \caption{XGBoost}
        \label{fig:violin_xgb}
    \end{subfigure}
    \caption{Violin plots displaying the distribution of the predictions for the three models trained using different label imputation methods.}
    \label{fig:violins}
\end{figure*}

\begin{figure}[]
    \centering
    \begin{subfigure}[b]{0.3\textwidth}
        \includegraphics[width=\textwidth]{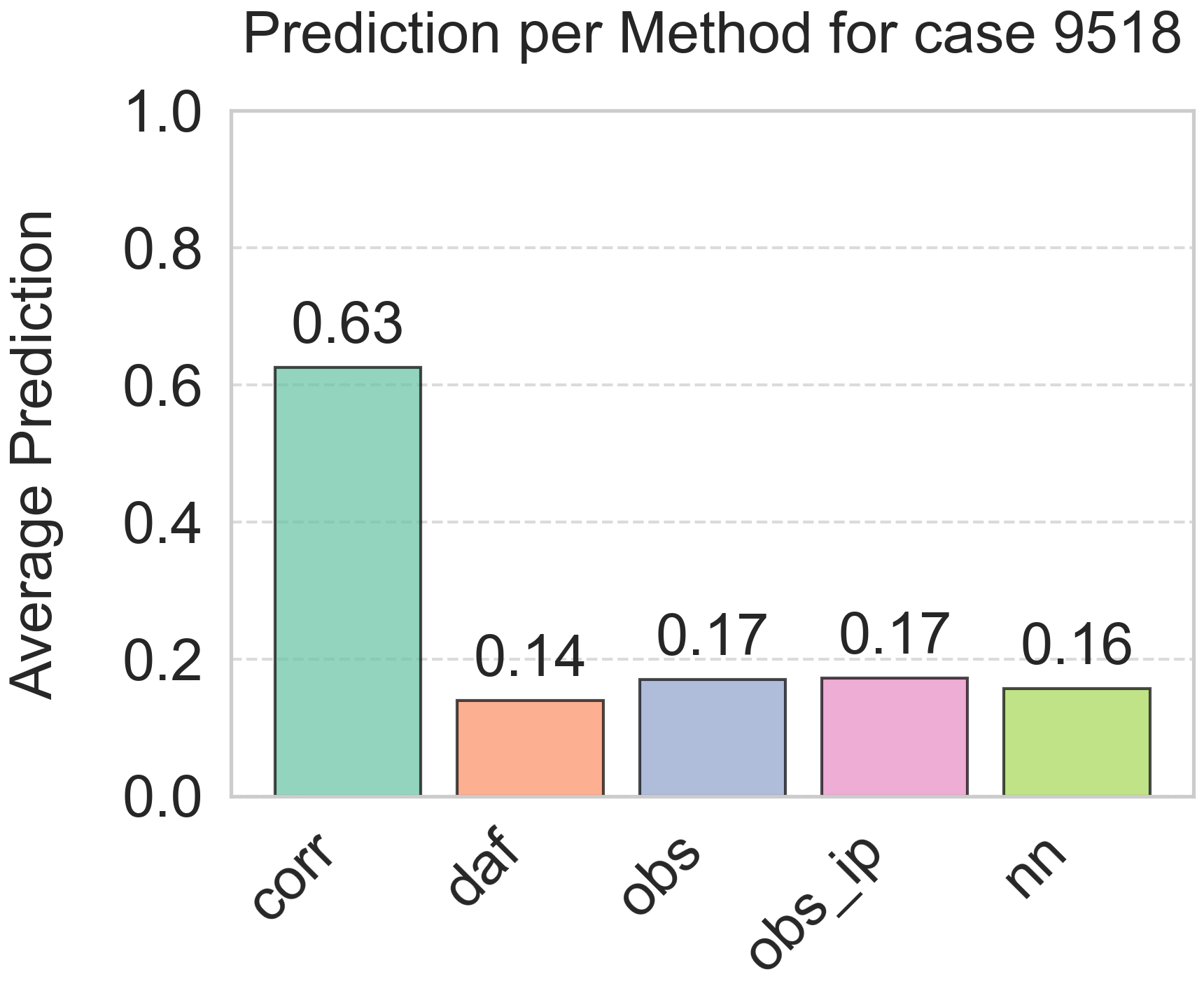}
        %\vspace{2pt}
        % \caption{Logistic Regression}
        % \label{fig:violin_lr}
    \end{subfigure}
    \vspace{10pt}
    \begin{subfigure}[b]{0.3\textwidth}
        \includegraphics[width=\textwidth]{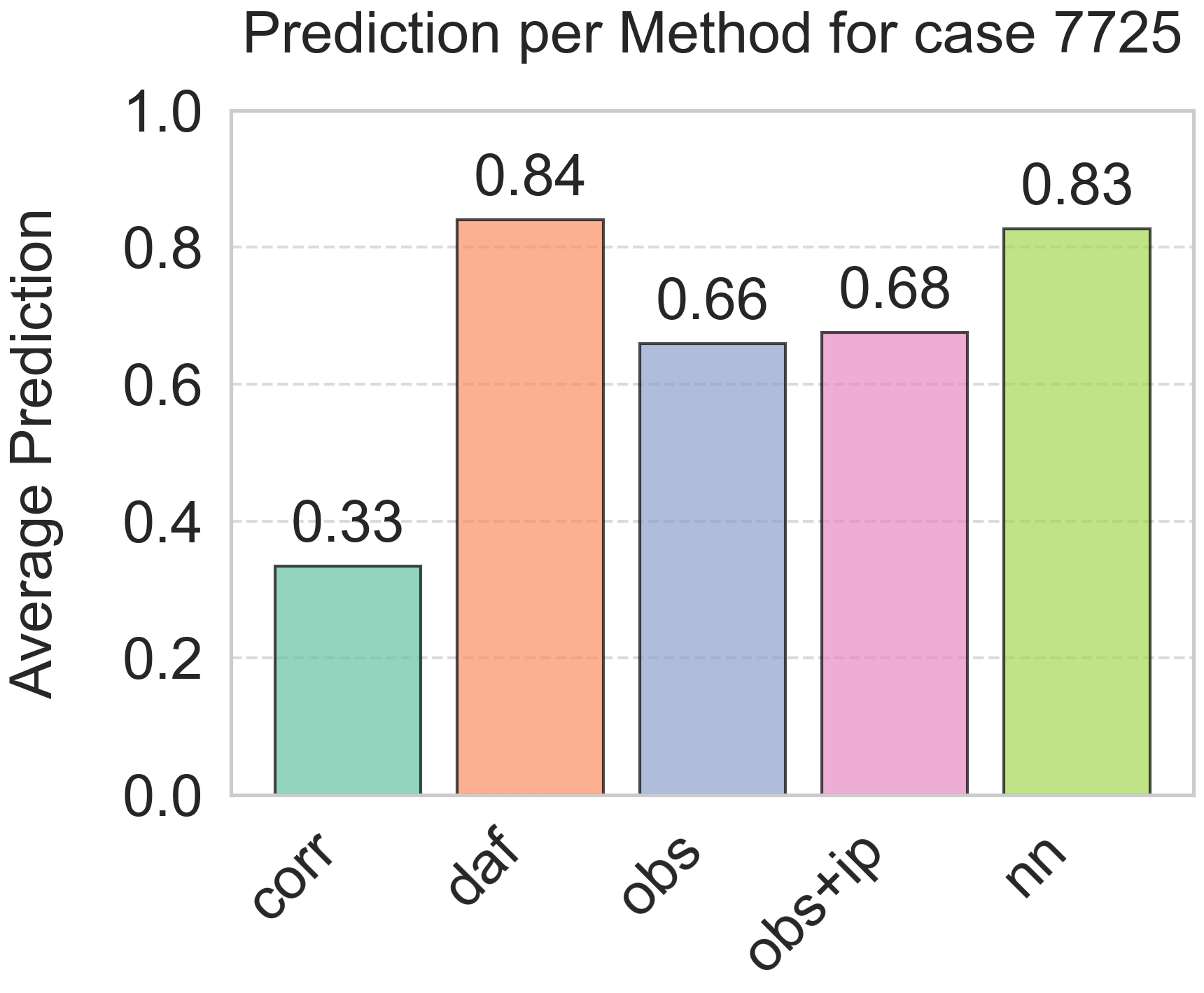}
        %\vspace{2pt}
        % \caption{Random Forest}
        % \label{fig:violin_rf}
    \end{subfigure}
    \caption{The mean prediction of the XGBoost model for two specific cases using different label imputation methods.}
    \label{fig:individual}
\end{figure}

\begin{figure*}
    \includegraphics[width=\textwidth]{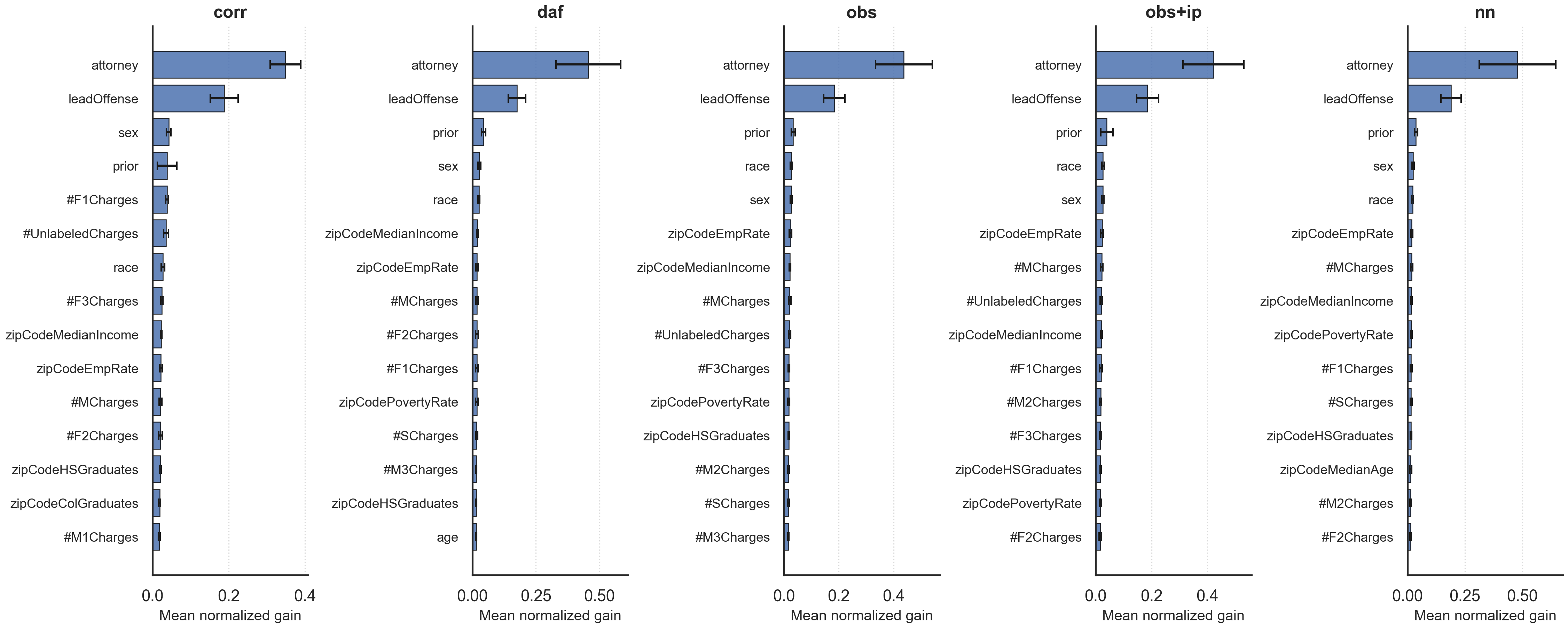}
    \caption{Top 15 most important features of the XGBoost model per label imputation method, ranked by mean normalized gain, averaged across the 25 training subsets.}
    \label{fig:xai}
\end{figure*}

In Table~\ref{tbl:performances}, we present model performance on the test set, separated by determinate and indeterminate cases. We report the mean Matthew's Correlation Coefficient, scaled from -100 to 100, along with the standard deviation. Note that labels for indeterminate cases are treated as unknown in this study, so model performance on these cases should be interpreted cautiously. Likewise, performance on determinate cases reflects only a restricted subset (cases where bail was granted) and may not generalize to the full dataset.

In Figure~\ref{fig:violins}, we show the prediction distribution of each of the three models trained using each of the five label imputation methods, illustrating the predictive behaviour of the model.

We zoom in on the predictions for two individual cases in Figure~\ref{fig:individual}, which shows the mean predicted FTA probabilities produced by the XGBoost model across the five label imputation methods.

Additionally, we investigate how the five label imputation methods affect the internal decision-making of the models. In Figure~\ref{fig:xai}, we report on the feature importance scores of the XGBoost classifier based on the gain metric. This metric reflects how much each feature contributes to improving the model’s predictions when it is used to split the data during training. For each label imputation method, normalized gain scores are averaged across the 25 training subsets. We report the top 15 features ranked by normalized gain score in Figure~\ref{fig:xai}.

\section{Discussion}
We first present the experimental results, then examine the label imputation methods from a legal-philosophical perspective, and finally discuss the study’s limitations and directions for future research.

\subsection{Discussion of results}
The violin plots in Figure~\ref{fig:violins} shows that the prediction distributions can vary quite significantly depending on the label imputation method used. The behaviour of the model is thus dependent on the way label indeterminacy is accounted for. This effect was also observed in the domain of overturned decisions in the European Court of Human Rights~\cite{StegingJURIX2025} and the medical domain~\cite{Schoeffer2025Perils}. The prediction distribution of the models trained using the $obs$ and $obs\_ip$ method seem quite similar however, which was also seen in previous research. While the $obs\_ip$ aims to correct the sample bias of the $obs$ method, seemingly no change can be observed in the resulting predictions of the models.

We confirmed these observations quantitatively using statistical tests (Wasserstein distance and Kolmogorov–Smirnov), which show that the differences between $obs$ and $obs_ip$ are negligible, whereas $corr$, $daf$, and $nn$ produce meaningfully distinct prediction distributions. While the violin plots in Figure~\ref{fig:violins} highlight differences in the overall shape and spread of predictions across models, the Wasserstein distances capture shifts in the distributions’ absolute values across methods. Overall, both visual and quantitative analyses indicate that the choice of label imputation method substantially shapes model behaviour. Furthermore, further statistical analysis and a direct comparison of Wasserstein distances across all methods demonstrates that label imputation induces larger changes in predictions than the choice of model type.

Figure~\ref{fig:individual} shows the average predictions of the XGBoost model on two specific cases, further illustrating how the choice of label imputation method can shape the output of the model. To provide insight into the models' decision-making, Figure~\ref{fig:xai} presents the most important features identified by the XGBoost model. Certain features appear consistently across imputation methods; for instance, the leading offense and type of attorney rank among the top three features for all five methods, highlighting their strong predictive value. Overall, nine features appear in the top 15 across all methods, though their relative importances vary. An additional twelve features are present in the top 15 for only three, two, or a single method. These variations demonstrate that the five methods differ fundamentally, not just in their predictions, but also in how they utilize the case information in the decision-making process.

In Table~\ref{tbl:performances}, we report model performance separately for determinate and indeterminate test cases. These results should be interpreted with caution, since labels for indeterminate cases are inherently uncertain. For instance, a model predicting failure to appear for a case in which the judge ordered pretrial detention would, counterintuitively, be treated as an incorrect prediction. While the determinate cases have an observed label (the defendant was granted bail and either appeared or failed to appear), this subset of cases is subject to sample bias, as it only contains cases in which bail was granted. Therefore, the performance scores alone do not convey the insights we typically expect.

What we can see in this Table, however, is that the $obs$, $obs_ip$, and $nn$ methods perform better on determinate cases than on indeterminate cases. In contrast, the $corr$ method performs worst on determinate cases but best on indeterminate cases. This pattern arises because $corr$ is the only approach that preserves the original labels of indeterminate cases. The other methods emphasize the observed determinate labels, explaining their stronger performance on determinate cases. This further illustrates how impactful the label imputation methods are on the predictive behaviour of the classifier.

\subsection{Legal perspective on label imputation}
% What methods for imputing the values of indeterminate labels makes most sense and why?

%TJZ: Do we want to put it here given the page limit or do we keep it for the journal paper? -> or maybe we dont care and try to push it as a long paper -> cut it afterwards?

Each of the five label imputation methods encodes a slightly different conception of justice in the context of risk governance. We argue that apart from purely technical consequences, design choices lead to normative commitments. The following subsection maps each method to its possible general legal-philosophical implications (See Table~\ref{tbl:methods} for method definitions and assumptions; see Figure~\ref{fig:bail_figure} for the role of counterfactuals in generating indeterminate labels). It should be emphasized that the following analysis does not aim to determine which of the examined imputation methods is normatively correct or preferable. Rather, we outline plausible lines of legal and philosophical arguments that may be advanced in relation to each approach. These arguments are not exhaustive, nor are they intended to be decisive. 

The \textit{corr} approach keeps the training data unchanged and includes denied-bail cases with their original recorded label, which is generally court appearances after detention. 
Defendants who were denied bail are assumed, for modeling purposes, to have appeared in court because they were detained, which gives rise to the `appearance' label. 
In doing so, the \textit{corr} method treats the resulting appearance labels as given and legitimate. 
Note that this might seem counter-intuitive, as the judge's decision to detain may intuitively equate to a `failure to appear' label (which is what we describe in the \textit{daf} method below). 
In the \textit{corr} method, however, appearance is taken as evidence of reliability and the detention is treated as an adequate safeguard against non-compliance, rather than as an intervention that alters behaviour. 
The model effectively treats the compelled appearance as equivalent to voluntary compliance, which is a doubtful assumption to say the least. For machine-learning models, this implies that \textit{corr} maps high-risk defendants, those who were denied bail and kept in pretrial detention, to `likely to appear in court'.  
One could argue that normatively, this reflects a preventive-justice perspective and that it potentially even encodes that restrictive measures are accepted as substitutes for individualized risk assessment \cite{koepke2018danger, McKay2020PredictingRisk, Zavrsnik2020AIHumanRights, AbuElyounes2020BailOrJail}. If followed, this reasoning shifts responsibility away from the state by allowing liberty-restricting interventions to produce the very outcomes that justify them. Under \textit{corr}, uncertainty about how defendants would have behaved if released is resolved by equating enforced compliance with genuine law-abiding conduct. As a result, the approach risks normalizing precautionary detention while hiding the underlying uncertainty. 

The \textit{daf} approach labels all indeterminate, pretrial-detention cases as FTA under the assumption that the prior judicial decision to detain correctly anticipated non-appearance. 
% In other words, \textit{daf} treats \textit{past decisions} as the proper basis for imputing counterfactual outcomes. 
% By contrast, the \textit{corr} approach assumes that \textit{observed labels} are correct (rather than that judicial decisions were justified): it simply records that detained defendants did in fact appear and therefore treats these compelled appearances as genuine evidence of low risk. %summarized below
Whereas \textit{corr} assumes that appearances due to pretrial detention reflect genuine compliance, \textit{daf} treats past detention decisions as indicators of FTA.
% This is of course a very doubtful assumption to say the least. For machine-learning models, this implies that \textit{corr} maps high-risk defendants, those denied bail, to no-FTA.  %Moved to above section
In legal-philosophical terms, under \textit{daf} detention decisions are read as reliable indicators of the possible FTA, which, by consequence legitimizes precautionary detention as a proof of risk and empowers epistemic authority of past decision-makers \cite{Stein2008EpistemicAuthority}. It is easy to notice the risk that this could lead to a dynamic of self-fulfilling-prophecy loop \cite{bauerself}: detention-based predictions are trained on labels derived from earlier detention decisions, which are taken as a source of evidence regarding the risk of FTA, reinforcing and amplifying initial risk assessments over time in a loop. In recividity prediction, this has been shown to perpuate harmful bias~\cite{COMPAS}. Both the \textit{corr} and \textit{daf} approaches are questionable for prediction purposes, but they can potentially serve as lower and upper bounds, e.g., in the context of sensitivity testing.

The \textit{obs} method limits modeling to cases where outcomes are observed, meaning cases in which bail was granted. Only appearances or failures to appear that occurred naturally, without pretrial detention, are treated as valid evidence. On the surface, \textit{obs} avoids speculating about what might have happened in indeterminate, pretrial detention cases. In practice, however, it leaves past discretionary decisions largely unexamined by treating indeterminate cases as missing data. As a result, \textit{obs} allows historical patterns of judicial discretion to persist. The model is trained on a selective sample and implicitly assumes that excluded cases are missing for neutral reasons~\cite{Zavrsnik2020AIHumanRights}. This assumption is questionable, as in fact, judges’ assessments of risk often shape which cases are granted bail. Under these conditions, \textit{obs} can reproduce and reinforce existing biases, while appearing neutral by placing responsibility on the data rather than on the modeling choices.

Both \textit{obs + ip} (inverse-propensity re-weighting) and \textit{nn} (nearest-neighbor imputation) attempt to address missing outcomes through statistical adjustment or substitution. They rely on informed approximations rather than direct legal judgments. Method \textit{obs+ip} re-weights observed cases to approximate the full population based on a specified propensity model, while \textit{nn} assigns outcomes to detained defendants by comparing them to similar released defendants. Those approaches depend on strong and often untestable assumptions, such as whether the available variables are sufficient for estimating propensities or whether the chosen similarity measure is meaningful. These assumptions are rarely fully verifiable. Consequently, one can also argue that the methods reflect a form of technocratic confidence in statistical tools to fill the gaps left by the legal process~\cite{McKay2020PredictingRisk}. They assume that modeling can reliably replace missing observations. This could pose a risk in which weak probabilistic inference is turned into decisive evidence regarding individuals behaviour and, as a consequence, their freedom. 

Furthermore, the statistical adjustment methods might pose a risk to individual contestability by affected individuals. An imputed FTA label is not a verifiable event. Defendants cannot readily challenge a counterfactual label, as they cannot demonstrate what would have occurred had bail been granted. Moreover, in practice there is a substantial risk that key properties of the model are not fully disclosed, such as how the data are preprocessed or which similarity metrics are used. At the same time, reliance on such models and different data imputation methods may redistribute responsibility across institutions, such as between judges and those developing or deploying a particular model~\cite{McKay2020PredictingRisk, Zavrsnik2020AIHumanRights}. As a result, there is a risk that crucial factors in decision-making may no longer be rebuttable by those who are directly and most deeply affected by them.

With that being said, even when statistically defensible, different imputation methods can carry significant philosophical and legal implications. Accordingly, while we do not seek to resolve these debates or argue for or against the use of such models, we demonstrate that technical design choices operate within, rather than outside of, broader legal and moral discourses.

\subsection{Limitations and future research}
The fundamental challenge of label indeterminacy lies in determining which labels should be considered indeterminate. In our case, we focus on predicting failure to appear and therefore treat labels as indeterminate when the observed failure to appear outcome is affected by an intervention, most notably pretrial detention.
Some additional questions arise regarding what should be considered indeterminacy. 
For example, when a judge sets monetary bail at a level that a defendant cannot afford, leading to pretrial detention, it is unclear whether this outcome should be treated as equivalent to a denial of bail and thus considered indeterminate, since one could argue that bail was technically granted. Defendants who are unable to pay monetary bail account for approximately one-third of the pretrial detention population in the United States~\cite{RANSON2023102872}, highlighting the practical importance of this issue. In our experiment, we treated cases in which bail was set but not paid as indeterminate, although the broader implications for how determinacy is defined warrant systematic investigation in future work.

Judges typically have access to bail reports containing substantially more information than is available in our dataset. Our models, by contrast, are limited to predicting failure-to-appear likelihoods and implicitly treat this as the sole basis for bail decisions, thereby excluding considerations such as public safety risk.
This limitation does not affect the conceptual focus of the study, however, as the objective is not to develop a state-of-the-art classifier, but to illustrate how label indeterminacy arises in bail decision-making.
For this reason, we focus on standard machine learning models that are relatively explainable and avoid more complex neural architectures. 
In practice, one could already argue that random forest and XGBoost models lack sufficient interpretability for such a highly sensitive task~\cite{Kavzoglu2022}. 
We therefore argue that symbolic or neuro-symbolic approaches may be more appropriate for this task, and that a human should always be kept in the loop.
Additionally, we aim to examine how label imputation methods influence fairness metrics in future work.
Finally, we did not include label imputation methods that require additional expert annotations and leave these for future research.

\section{Conclusion}
This paper examined the role of label indeterminacy in machine learning models for predicting failure to appear in pretrial bail decisions. A substantial portion of historical cases do not reliably encode defendants’ appearance behaviour, as outcomes are often shaped by interventions such as pretrial detention. Treating these cases as fully informative introduces implicit and unverifiable assumptions that materially affect model behaviour.

Our empirical evaluation of multiple approaches to handling indeterminate labels across several machine learning models showed that methodological choices surrounding label treatment can have as much, or more, impact than model selection itself. Explainable AI analysis further revealed that these choices alter the internal decision-making of models, not just predictive performance. 
Moreover, each label imputation method is based on a set of unverifiable assumptions. We discuss how these assumptions carry subtle normative and legal implications and map each method to legal and philosophical arguments, highlighting how technical design choices operate within broader moral and legal discourses.
These findings highlight label indeterminacy as a central concern with direct consequences for fairness, performance, and accountability in predictive models used for technological bail decision support systems.

\begin{acks}
This research was funded by the Hybrid Intelligence Center, a 10-year programme funded by the Dutch Ministry of Education, Culture and Science through the Netherlands Organisation for Scientific Research, https://hybrid-intelligence-centre.nl. The research leading to these results has received funding from the National Science Centre, Poland, project no. 2025/57/N/HS5/01561, titled “Uncertainty and Argumentation:
Decision-Making Under Uncertainty in Legal Disputes.”
\end{acks}

\bibliographystyle{ACM-Reference-Format}
\bibliography{sample-base}

% \clearpage
\appendix
\section*{Appendix}

\section{Model hyperparameters}
The full parameters of each model used in this study can be seen in Table~\ref{tbl:appendix_parameters}. We include these for transparency and reproducibility reasons. Most settings were kept at default, or were inspired by previous research~\cite{Schoeffer2025Perils, Han2021RecidivismXGBoost}. The full code will be made available upon acceptance.

\begin{table}[hb!]
\centering
\caption{Hyperparameters used for each model}
\begin{tabular}{ll}
\toprule
\multicolumn{2}{l}{\textbf{Logistic Regression}} \\
\midrule
Maximum iterations  & 2000 \\
Regularization & $\ell_2$ \\
\midrule
\multicolumn{2}{l}{\textbf{Random Forest}} \\
\midrule
Number of trees & 500 \\
Maximum tree depth & 8 \\
Minimum samples per leaf  & 50 \\
Feature subsampling  & $\sqrt{p}$ \\
\midrule
\multicolumn{2}{l}{\textbf{XGBoost}} \\
\midrule
Objective & binary:logistic \\
Number of trees & 800 \\
Maximum tree depth & 4 \\
Learning rate ($\eta$) & 0.005 \\
Row subsampling  & 0.8 \\
Column subsampling & 0.9 \\
$\ell_2$ regularization ($\lambda$) & 5.0 \\
$\ell_1$ regularization ($\alpha$) & 1.0 \\
Minimum child weight & 20 \\
\bottomrule
\end{tabular}
\label{tbl:appendix_parameters}
\end{table}

\end{document}